# Improving Automatic Skin Lesion Segmentation using Adversarial Learning based Data Augmentation

Lei Bi[1], Dagan Feng[1, 2] and Jinman Kim[1]

[1] BMIT Group, School of Information Technologies, the University of Sydney, Australia
[2] Med-X Research Institute, Shanghai Jiao Tong University, China
{lei.bi, dagan.feng, jinman.kim}@sydney.edu.au

## Abstract

Segmentation of skin lesions is considered as an important step in computer aided diagnosis (CAD) for automated melanoma diagnosis. In recent years, segmentation methods based on fully convolutional networks (FCN) have achieved great success in general images. This success is primarily due to the leveraging of large labelled datasets to learn features that correspond to the shallow appearance as well as the deep semantics of the images. However, the dependence on large dataset does not translate well into medical images. To improve the FCN performance for skin lesion segmentations, researchers attempted to use specific cost functions or add post-processing algorithms to refine the coarse boundaries of the FCN results. However, the performance of these methods is heavily reliant on the tuning of many parameters and post-processing techniques. In this paper, we leverage the state-of-the-art image feature learning method of generative adversarial network (GAN) for its inherent ability to produce consistent and realistic image features by using deep neural networks and adversarial learning concept. We improve upon GAN such that skin lesion features can be learned at different level of complexities, in a controlled manner. The outputs from our method is then augmented to the existing FCN training data, thus increasing the overall feature diversity. We evaluated our method on the ISIC 2018 skin lesion segmentation challenge dataset and showed that it was more accurate and robust when compared to the existing skin lesion segmentation methods.

## 1    Introduction

Malignant melanoma has one of the most rapidly increasing incidences in the world and has a considerable mortality rate. Early diagnosis is particularly important since melanoma can be cured with prompt excision. Dermoscopy images play an important role in the non-invasive early detection of melanoma [1]. However, melanoma detection using human vision alone can be subjective, inaccurate and poorly reproducible even among experienced dermatologists [2]. This is attributed to the challenges in interpreting images with diverse characteristics including lesions of varying sizes and shapes, lesions that may have fuzzy boundaries, different skin colors and the presence of hair [2]. Motivated by these difficulties, there has been a great interest in developing computer-aided diagnosis (CAD) systems that can assist the dermatologists' clinical evaluation [1, 2].

Segmentation of skin lesions is considered as an important step for a melanoma CAD. However, traditional methods [3, 4] that use edges, regions and shape models, depend heavily on hand-crafted features and priori knowledges, which inhibit widespread application.

Recently, deep learning methods based on fully convolutional networks (FCN) have achieved great success in natural image segmentation related challenges [5]. This success is primarily attributed to the ability of FCN to leverage large datasets to hierarchically learn the features that best correspond to the appearance as well as the semantics of the images [5]. In addition, FCN can be trained in an end-to-end manner for efficient inference, i.e., images are taken as inputs and the segmentation results are directly outputted. However, there is a scarcity of annotated medical images training data due to the large cost and complicated acquisition procedures [6]. Consequently, without sufficient training data to cover all the variations, where for instance, skin lesions from different patients can have major differences in textures/size/shape, FCN fails to provide accurate results. Although data augmentation approaches, such as random crops, flips and color jittering, are often applied to increase the overall volume of the training data, they are simply duplicating the existing training features, which thereby is not capable of producing a variety of new features for leaning.

To improve the FCN performance for skin lesion segmentations, researchers attempted to use specific cost functions or add post-processing algorithms to refine the coarse boundaries of the FCN results. For example, Yuan et al [7] replaced the cross-entropy loss used in traditional FCN with a Jaccard distance loss for training. Bi et al [8] used cellular automata algorithm as a post-processing algorithm to refine the FCN segmentation outcomes. Unfortunately, data specific cost functions have limited generalizability to different datasets. In addition, the reliance on post-processing algorithms could override the FCN outcomes, which is due to the fact that the post-processing is usually unsupervised and cannot fully described the training data.

### 1.1 Our Contributions

In this paper, we propose to improve the segmentation performance of FCN by a novel adversarial learning approach. We leverage the state-of-the-art image feature learning method of generative adversarial networks [9] (GAN) for its inherent ability to produce consistent and realistic image features by using deep neural networks and adversarial learning concept. We improve upon GAN to learn skin lesion features at different levels of complexities, in a controlled manner, and then added the learned skin lesion features into the existing FCN training data, which thereby increase the overall feature diversity. The ability to improve the overall feature diversity enables the FCN to learn from greater image feature variations, in an iterative manner, to improve the segmentation accuracy. In addition, our method is generalizable to any FCN architectures.

## 2      Methods and Materials

### 2.1      Materials

ISIC 2018 [10, 11] is a subset of the large International Skin Imaging Collaboration (ISIC) archive, which contains dermoscopic images acquired on a variety of different devices at numerous leading international clinical centers. The ISIC 2018 skin lesion segmentation challenge dataset provides 2,594 training images. Image size varied from 540×576 pixels to 4499×6748 pixels. Clinical experts provided manual delineations in the training data.

For evaluation purpose, we further randomly split the training images into 2,335 images for training and 259 images for training validation. Additional 100 images were provided by the challenge organizer as the test validation set. The ground truth for the test validation set is not available to the public and the results were processed by the online submission system.

### 2.2      Fully Convolutional Networks

The FCN architecture was converted from convolutional neural networks (CNNs) for efficient dense inference [5]. It contains downsampling and upsampling parts. The downsampling part has stacked convolutional layers to extract high-level semantic information and has been routinely used in CNNs for image classification related tasks [12]. The upsampling part has stacked deconvolutional, which are transposed convolutional layers that upsample the feature maps derived from downsampling part to output the segmentation results. For skin lesion segmentation, the FCN architecture can be trained end-to-end by minimizing the overall loss function (e.g., cross-entropy loss) between the predicted results and the ground truth annotation of the training data. The FCN parameters (weights) can then be updated iteratively using e.g., stochastic gradient descent (SGD) algorithm.

### 2.3      Adversarial Learning for Skin Lesion Features

Adversarial Learning (also known as generative adversarial networks (GAN) [9]) has 2 main components: a generative model $G$ (the generator) that captures the data distribution and a discriminative model $D$ (the discriminator) that estimates the probability of a sample that came from the training data rather than $G$. The generator is trained to produce outputs that cannot be distinguished from the real data by the adversarially trained discriminator, while the discriminator was trained to detect the synthetic data created by the generator.

For learning the skin lesion features, we embed the training label (annotation) for training and adoption as part of the formulation. During training, the generator takes the training label as the input to learn a mapping to synthesize the dermoscopy images that appear realistic. The discriminator then attempts to separate the real and

Submitted to the Arxiv.org on the July 23rd, 2018

synthesized dermoscopic images. Thus the loss function can be defined as conditional [13, 14] on the label $l$:

$$\mathcal{L}(G,D) = \mathbb{E}_{l,y}[logD(l,y)] + \mathbb{E}_{l,z}[\log(1 - D(l, G(l,z)))]$$

where $y$ is the dermoscopy images and $z$ is the input random noise. $D(\cdot)$ represents the probability that the input to $D(\cdot)$ came from the real data while $G(\cdot)$ represents the mapping to synthesize the real data.

### 2.4 Adversarial Learning for Improving Segmentation

Initially, we used the training data (denote as $R$), including the ground truth label (denote as $R_l$) and images (denote as $R_y$) to train theFCN model. The trained FCN model was then applied on $R$ to separate the data into two equal partitions. Based on the segmentation performance, dice similarity coefficient, we equally separated the $R$ into a simple (denote as $S$) and complex (denote as $C$) training sets. Afterwards, we used the label $S_l$ and images $S_y$ in $S$ to train a S-Model based on the adversarial learning for deriving ROI features. The same approach was also used to train a C-Model. At the adoption stage, C-Model was applied on the label $S_l$ to produce the additional complex training data $C_y^*$ while S-Model was applied on the label $C_l$ to get the additional simple training data $S_y^*$. Finally, the original training data $R$ together with the derived additional training data were used to train a new FCN for skin lesion segmentation. In this work, we derived additional 2,335 images from adversarial learning for training the FCN.

## 3 Experiments and Results

The evaluation was conducted on both the train validation and test validation sets (See Section 2.1 for more details).

For evaluation, we compared the inclusion of adversarial learning (AL) for segmentation. The more recent FCN using residual network architecture (101-layer, denote as ResNet [15-18]) was used. To further improve the segmentation performance, an ensemble approach was used (denoted as ENS). More specifically, the final output was produced by integrating the segmentation outputs derived from top 5 trained models.

For the training validation set, the common segmentation evaluation metrics were used: dice similarity coefficient (Dice), Jaccard (Jac.), sensitivity (Sen.), specificity (Spe.) and Accuracy (Acc.). For the test validation set, the threshold Jaccard index (T-Jac.) was used by the online submission system, where the image with less than 0.65 Jac will be given a value of 0.

**Table 1.** Segmentation results for the train validation dataset.

| %             | Dice  | Jac.  | Sen.  | Spe.  | Acc.  |
|---------------|-------|-------|-------|-------|-------|
| **ResNet**        | 88.82 | 81.86 | 91.64 | 95.94 | 95.90 |
| **ResNet-AL**     | 89.26 | 82.40 | **93.34** | 95.99 | 95.92 |
| **ResNet-AL-ENS** | **89.79** | **83.12** | 92.49 | **96.65** | **96.23** |

**Table 2.** Segmentation results for the test validation dataset.

| %             | T-Jac. |
|---------------|--------|
| **ResNet**        | 74.30  |
| **ResNet-AL**     | 74.80  |
| **ResNet-AL-ENS** | **77.80** |

Table 1 and Table 2 show that our method can improve FCN based segmentation with a much higher Jaccard index. We attribute this improvement to the adversarial learning method for deriving additional features for training which thereby improves the overall feature diversity and the skin lesion detection performance.